%% file: main.tex
\documentclass[letterpaper, 10 pt, conference]{ieeeconf}  %

\IEEEoverridecommandlockouts                              %

\title{\LARGE \bf
Risk-Aware Off-Road Navigation via a Learned Speed Distribution Map
}

\author{Xiaoyi Cai$^\dagger$, Michael Everett$^\dagger$, 
Jonathan Fink$^\ddagger$, and Jonathan P. How$^\dagger$%
\thanks{$^\dagger$Dept. of Aeronautics and Astronautics, Massachusetts Institute of Technology, Cambridge, MA 02139, USA. {\tt\{xyc, mfe, jhow\}@mit.edu}.}%
\thanks{$^\ddagger$U.S. Army Combat Capabilities Development Command Army Research Laboratory, Adelphi, MD 20783, USA. {\tt jonathan.r.fink3.civ@army.mil}.}%
}

\input{macros}

\begin{document}

\maketitle
\thispagestyle{empty}
\pagestyle{empty}

\begin{abstract}
Motion planning in off-road environments requires reasoning about both the geometry and semantics of the scene (e.g., a robot may be able to drive through soft bushes but not a fallen log). 
In many recent works, the world is classified into a finite number of semantic categories that often are not sufficient to capture the ability (i.e., the speed) with which a robot can traverse off-road terrain.
Instead, this work proposes a new representation of traversability based exclusively on robot speed that can be learned from data, offers interpretability and intuitive tuning, and can be easily integrated with a variety of planning paradigms in the form of a costmap.
Specifically, given a dataset of experienced trajectories, the proposed algorithm learns to predict a distribution of speeds the robot could achieve, conditioned on the environment semantics and commanded speed.
The learned speed distribution map is converted into costmaps with a risk-aware cost term based on conditional value at risk (CVaR).
Numerical simulations demonstrate that the proposed risk-aware planning algorithm leads to faster average time-to-goals compared to a method that only considers expected behavior, and the planner can be tuned for slightly slower, but less variable behavior.
Furthermore, the approach is integrated into a full autonomy stack and demonstrated in a high-fidelity Unity environment and is shown to provide a 30\% improvement in the success rate of navigation.
\end{abstract}

\input{intro}

\input{related_work}
\input{traversability}
\input{min_time_navigation}
\input{results}
\input{conclusion}

\section*{Acknowledgment}
Research was sponsored by the Army Research Office and was accomplished under Cooperative Agreement Number W911NF-21-2-0150. The views and conclusions contained in this document are those of the authors and should not be interpreted as representing the official policies, either expressed or implied, of the Army Research Office or the U.S. Government. The U.S. Government is authorized to reproduce and distribute reprints for Government purposes notwithstanding any copyright notation herein.

\bibliographystyle{IEEEtran}
\bibliography{bibs}

\end{document}

%% file: macros.tex
\usepackage{amsmath,amsthm,amssymb, amsfonts} %
\usepackage{bbold}
\usepackage{mathtools}
\usepackage{graphicx, xcolor}
\usepackage{pdfpages}
\usepackage{bm}
\usepackage{url}
\usepackage{enumerate}
\usepackage{float}
\usepackage[sort, compress]{cite}
\usepackage{cases,balance}
\usepackage[pdftex, pdfstartview={FitV}, pdfpagelayout={TwoColumnLeft},bookmarksopen=true,plainpages = false, colorlinks=true, linkcolor=black, citecolor = black, urlcolor = black,filecolor=black , pagebackref=false,hypertexnames=false, plainpages=false, pdfpagelabels ]{hyperref}

\usepackage[capitalise]{cleveref}

\usepackage[font=footnotesize]{caption}
\usepackage{subcaption}

\usepackage{array}
\usepackage{booktabs}
\usepackage{multirow} %

\usepackage{comment} %
\excludecomment{comments} %

\usepackage{algorithm}
\usepackage[noend]{algpseudocode} %
\definecolor{commentclr}{RGB}{110, 149, 204}

\makeatletter
\newcommand\fs@spaceruled{\def\@fs@cfont{\bfseries}\let\@fs@capt\floatc@ruled
  \def\@fs@pre{\vspace{0.6\baselineskip}\hrule height.8pt depth0pt \kern2pt}%
  \def\@fs@post{\kern2pt\hrule\relax}%
  \def\@fs@mid{\kern2pt\hrule\kern2pt}%
  \let\@fs@iftopcapt\iftrue}
\makeatother

\usepackage{tikz}

\newcommand{\scmd}{s^{\text{cmd}}}

\newcommand{\setS}{S} %
\newcommand{\setM}{\mathbf{M}} %
\newcommand{\setSM}{\mathbf{M_s}} %
\newcommand{\setO}{O} %

\newcommand{\smax}{s^{\text{max}}} %

\newcommand{\done}[1]{\mathbb{1}^{\text{done}}(#1)} %

\newcommand{\slookup}[1]{\mathcal{S}(#1)}

\newcommand{\cvar}[2]{{\text{CVaR}_{#1}(#2)}}
\newcommand{\var}[2]{{\text{VaR}_{#1}(#2)}}

\newcommand\norm[1]{\left\lVert#1\right\rVert}

\newcommand{\R}{\mathbb{R}}

\newcommand{\ie}{i.e.}

\newcommand*{\defeq}{:=}

\newcommand{\code}[1]{\texttt{#1}}

\theoremstyle{plain}%

\theoremstyle{remark}

\theoremstyle{definition}

\graphicspath{{Figs/}}

%% file: intro.tex
\section{Introduction}

\begin{figure}[t]
	\centering
	\includegraphics[width=0.9\linewidth, trim={0cm 0cm 0cm 0cm},clip]{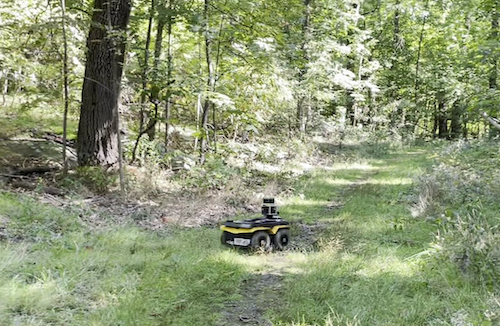}
	\caption{
		Off-Road Navigation.
		Even though much of this scene can be categorized broadly as grass/foliage, the vehicle's ability to traverse through different regions can vary dramatically.
		This paper proposes a planner that accounts for such variability within semantic terrain/object classes, by learning a speed distribution map.
	}
	\label{fig:jackal_offroad}
\end{figure}

Autonomous robotic navigation in off-road environments is important for many tasks, including planetary exploration~\cite{bares1989ambler,massari2004autonomous} and search-and-rescue missions~\cite{kantor2003distributed}.
In an off-road setting, a fundamental challenge in the local planning problem is that both the \textit{geometry} (e.g., positive and negative obstacles) and the \textit{semantics} (e.g., terrain type, time of year) impact the speed and safety with which a robot can traverse the environment.
For example, \cref{fig:jackal_offroad} shows a mobile robot surrounded by grass and foliage.
While a purely geometry-based system might consider much of the scene as obstacles (e.g., since a lidar pointcloud would have many returns from the leaves and grass), the semantics suggest much of the vegetation is sufficiently soft that the vehicle could drive through it, but trees and rocks would likely stop the robot.

To capture these semantics, many approaches train semantic segmentation modules for camera images~\cite{valada16iser,valada2017adapnet} or lidar pointclouds~\cite{shaban2022semantic}, which can reduce the dimensionality and enable learning in a low visual fidelity simulation environment~\cite{sadeghi2016cad2rl,everett2019planning}.
However, the ontology of existing labeled datasets for off-road navigation~\cite{valada16iser,RUGD2019IROS,jiang2021rellis,maturana2018real} is a fundamental limitation for capturing traversability.
For instance, the 20 and 24 classes in \cite{RUGD2019IROS,jiang2021rellis} only contain the broad ``bush'', ``grass'', and ``tree'' classes for vegetation, where the varying degree of traversability within each class is not captured. For instance, some bushes can be driven through, some will slow the robot down, and others will stop the robot.
Alternatively, \cite{maturana2018real} contains labels with finer granularity such as ``traversable grass'' and ``non-traversable low vegetation'', but these are specific to the large vehicle in mind during the expensive manual labeling procedure, and thus would not generalize to other vehicles.

Furthermore, because each individual mission or operator can have different objectives, it is important that the risk tolerance of the planner (e.g., whether to take a shortcut that has a 1\% chance of causing the robot to get stuck) can be quickly adjusted.
Risk-aware planning has been extensively studied, e.g.,~\cite{ono2015risk,pereira2013risk,de2011minimum, blackmore2011chance, luders2010chance}, with different notions of risk such as collision probability and classification uncertainty. Recent works such as~\cite{hakobyan2019risk, fan2021step} have adopted conditional value at risk (CVaR) as the risk metric, which has been analyzed in-depth in~\cite{majumdar2020should} about why CVaR allows robots to assess risks rationally.
Alternatively, other work tries to directly learn policies, dynamics models, or cost functions~\cite{silver2010applied, silver2008high, bojarski2016end, pan2017agile, wang2021apple,kahn2021badgr, manderson2020learning} that satisfy the desired risk tolerance.
However, these off-road techniques often require a non-intuitive cost function tuning procedure or a complete re-training of the model to adjust the risk tolerance.

To address these issues and bridge the gap between semantic perception and risk-aware planning, this work proposes a new representation of traversability as a general distribution of robot speed conditioned on environment semantics and the commanded speed.
The proposed pipeline first automatically labels a dataset of collected trajectories with realized vehicle speeds to capture the variability in speed outcomes associated with each semantic class.
This dataset can then be used to learn a distribution of achievable speeds, conditioned on the commanded speed and semantics of the nearby terrain.
The learned speed distribution is then converted into a \textit{speed map} representation that can be leveraged with various planning paradigms.
Notably, we incorporate risk-awareness into the planner via CVaR and show how to adjust the risk without collecting any extra data or re-training the learned model.

The contributions of this work include:
i) a new representation of traversability as a probability distribution of speeds the robot could achieve, which can be learned from data and allows physically meaningful interpretation in m/s;
ii) a new risk-aware minimum-time planner based on Model Predictive Path Integral (MPPI~\cite{williams2017information}) control that uses the learned speed distribution map, allowing risk level adjustment without re-training or collecting more trajectories;
iii) a demonstration of a robot reaching its goal with up to $30$\% improvement in success rate than a risk-unaware algorithm in a high-fidelity Unity simulation environment in a full autonomy stack.

%% file: related_work.tex
\section{Related Work}

Traversability analysis can be achieved via both proprioceptive and exteroceptive~\cite{papadakis2013terrain} sensors, where the former category includes IMUs that measure vibration and orientation of the robot~\cite{oliveira2021three, otte2016recurrent}, and the latter category that includes lidar and RGB cameras that provide geometric and semantic understanding of the environment. 
Purely geometry-based analysis has been widely adopted, e.g.,~\cite{larson2011off, overbye2020fast, overbye2021g, fan2021step}, which often involves a weighted sum of costs extracted based on geometric properties such as slope, roughness and step height. Notably, geometry-based methodology has been demonstrated successfully in the DARPA Subterranean Challenge~\cite{fan2021step}, where the cost function is assumed to be Gaussian, allowing easy computation and adjustment of risk level via CVaR.
In contrast, this work proposes a new representation of traversability based on experience and brings semantics into the problem.

Among methods that combine semantic-based and geometry-based techniques, \cite{guan2021ttm} proposes a fusion strategy that uses geometry-based cost (slope added to step height) for the terrain, unless the associated semantic label is known to be undesirable. \cite{tan2021risk} uses both geometric and semantic layers in a multi-layer costmap, and fuses the costs by accounting for layer uncertainty. \cite{shaban2022semantic} classifies a dense 3D pointcloud to extract traversability labels (\texttt{Free}, \texttt{Low Cost}, \texttt{Medium Cost}, \texttt{Lethal}), where the ground truth labels are designed based on human expertise. 
These methods either require human expertise in associating semantics with traversability, or require combining costs with different units, which makes tuning non-intuitive.
In contrast, this work uses vehicle speed as a common unit in the cost function to enable intuitive risk level adjustment.

Other recent work proposes methods to learn navigation policies or cost functions from experience via imitation learning~\cite{silver2010applied, silver2008high, bojarski2016end, pan2017agile}, inverse reinforcement learning~\cite{wang2021apple}, semi-supervised learning~\cite{suger2015traversability}, or model-free reinforcement learning~\cite{wiberg2021control}.
While these methods leverage datasets or simulators to reduce some of the expert knowledge requirements, a key limitation is that adjusting the risk tolerance could require collection of a new set of expert trajectories and/or re-training the learned models.
Alternatively, \cite{kahn2021badgr, manderson2020learning} learn a predictive ``events'' (e.g., bumpy, collision, smooth) model from a diverse dataset of experiences. By predicting the probability of undesirable events, the riskiness of the planner in \cite{kahn2021badgr} can be adjusted by changing the penalty for these events without re-training the network. However, the cost terms in~\cite{kahn2021badgr} have different units such as terrain bumpiness and goal proximity, leading to a difficult conversion from risk tolerance to reward function weights.
This work also leverages learning from a dataset of experienced trajectories, but instead proposes a pipeline to produce speed maps that can be incorporated into many planning paradigms.

%% file: traversability.tex
\section{Traversability}
For fast off-road navigation, time-to-goal is a typical performance measure that depends on the quality of planned trajectory and how well a robot executes the planned maneuvers. A good assessment of terrain traversability allows a planner to generate trajectories that are both fast and can be executed. To this end, the discrepancy between the vehicle's \textit{planned} speed and \textit{realized} speed provides a natural quantity to describe the \textit{traversability} of terrain.
Importantly, traversability is probabilistic in nature, due to imperfect sensing, broad semantic class labels for terrain, and the dynamics of vehicle-terrain interactions. Therefore, we propose to capture traversability via a conditional distribution of realized speed given the commanded speed and sensor observation.

\subsection{Traversability as a Conditional Speed Distribution}
\label{sec:traversability_as_distribution}

Denoting the set of realized speeds as $\setS$ and the set of possible observations about a terrain patch as $\setO$, we define traversability of the terrain as the conditional distribution 
\begin{equation}
p_\theta(s \mid \scmd, o):\setS \mid \setS \times \setO \rightarrow \R, \label{eq:cond_distribution}
\end{equation}
where $s, \scmd \in\setS$ are the realized speed and the commanded speed, $o\in\setO$ is the observation about the terrain, and $p_\theta$ is a probability distribution parameterized by $\theta$, which in practice can be learned via a neural network. This representation is general enough to capture the multi-modality of the distribution and allows the planner to extract desired statistics for trajectory planning such as mean, modes and variance.

\begin{figure}[t]
	\centering
	\includegraphics[width=0.7\linewidth, trim={0cm 0cm 0cm 0cm},clip]{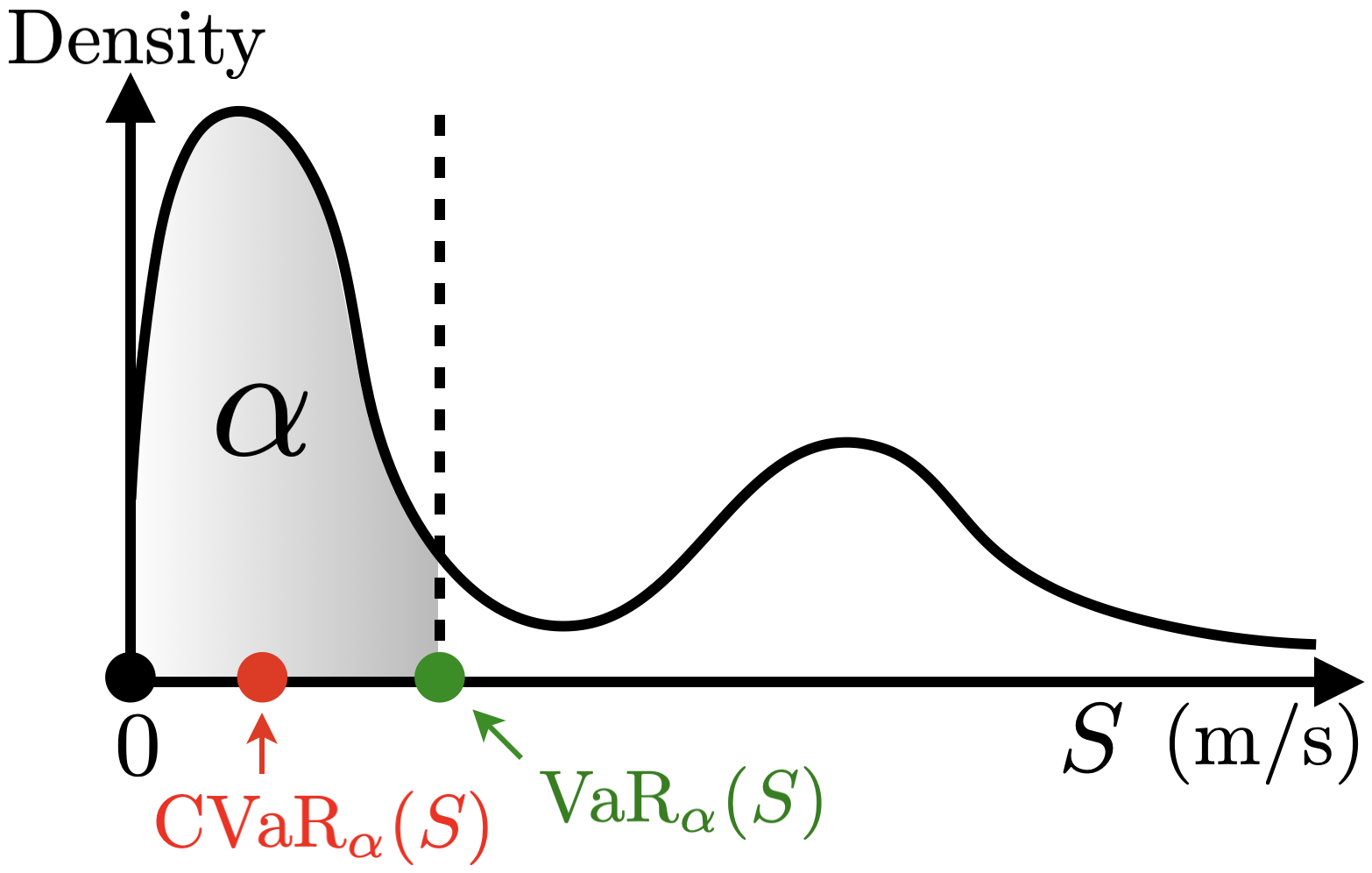}
	\caption{
	    This work defines conditional value at risk (CVaR) to capture worst case \textit{speed} (i.e., near $0$), rather than other common definitions based on the (right-side) tail.
	}
	\label{fig:cvar}
\end{figure}

\begin{figure*}[t]
	\centering
	\includegraphics[width=\linewidth, trim={0cm 0cm 0cm 0cm},clip]{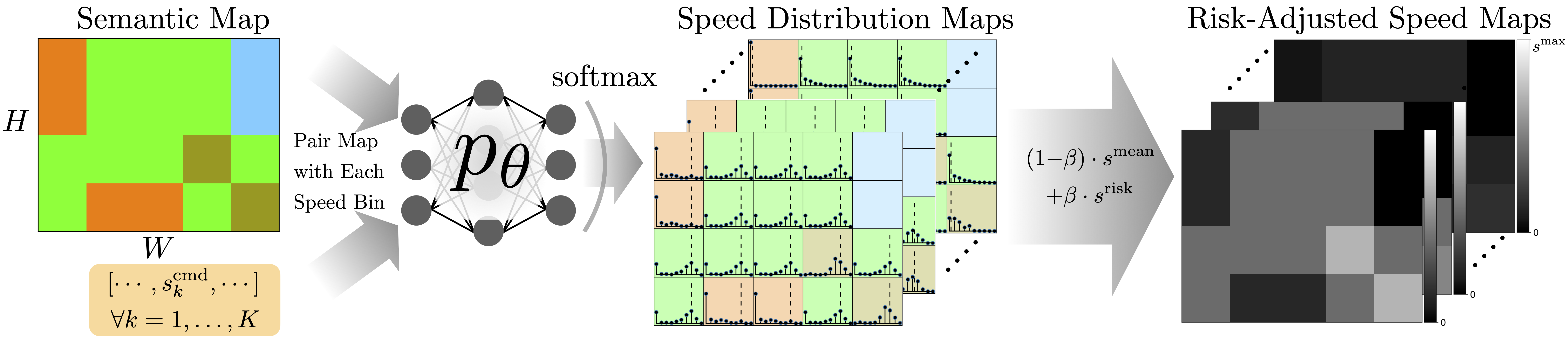}
	\caption{
		 Low-dimensional visualization of the pipeline to generate risk-aware speed map from semantic input and desired commanded speeds. A $10$-layer $100\times100$ map with $10$ output speed bins can be efficiently evaluated on an $8$-core CPU under $2$~ms with a simple neural network with two 64-node hidden layers. The speed distribution maps allow any planner to easily adjust the risk tolerance via relative weight $\beta$ or via the $\alpha$ level of CVaR directly.
	}
	\label{fig:architecture}
\end{figure*}

For high-speed navigation in a cluttered environment (e.g., a forest), the planner often has to trade off opportunities to reduce navigation time with the risk of colliding with an obstacle or getting stuck. To quantify these risks, we adopt the Conditional Value at Risk (CVaR), which satisfies a group of axioms important for rational risk assessment~\cite{majumdar2020should}.
The Conditional Value at Risk at level $\alpha\in[0,1]$ is defined as:
\begin{equation} \label{eq:cvar}
\cvar{\alpha}{\setS}\defeq \frac{1}{\alpha} \int_{0}^{\alpha} \var{\tau}{\setS}\ d\tau,
\end{equation}
where $\var{\alpha}{\setS}$ is the Value at Risk, or the $\alpha$-quantile:
\begin{equation} \label{eq:var}
\var{\alpha}{\setS}\defeq \max \{ s \mid p(\setS < s) \leq \alpha\}.
\end{equation}
Note that we define CVaR to capture the \textit{worst-case speed} outcomes (\ie, lowest speed), as visualized in Fig.~\ref{fig:cvar}.
Intuitively, $\cvar{\alpha}{\setS}$ measures the average speed outcomes that are lower than the $\alpha$-quantile of the speed distribution, capturing the worst-case expected speed. Notice that CVaR is the same as the mean when $\alpha=1$, so often a low $\alpha$ is picked for sufficient distinction from the expectation.

\subsection{Generating a Risk-Aware Traversability Map}

Next, we show how a learned speed distribution from~\cref{eq:cond_distribution} can be used to convert a semantic map (built from the robot's sensor data) into a representation of traversability for the planner.

The architecture is illustrated in~\cref{fig:architecture}.
The input to the pipeline is a semantic gridmap, $\setSM\in \R^{C \times H \times W}$, with $C$ semantic classes, width $W$, and height $H$.
Let $\setM\in \R^{K\times H\times W}$ represent a  $K$-layer speed map, where each layer has width $W$ and height $H$. For map $\setM$, we denote $m_{k,h,w}$ as the cell value in layer $k$, row $h$ and column $w$. 
Given the speed limits of $[0,\smax]$, we let the $k$-th layer correspond to the commanded speed range of $[\frac{(k-1)}{K}\smax, \frac{k}{K}\smax]$, where $k\in\{1,\dots,K\}$. Lastly, we associate the distribution $p_\theta(s\mid\scmd=\frac{(k-0.5)}{K}\smax, o_{h, w})$ to each cell indexed by $(k, h, w)$, where $o_{h, w}$ denotes the observation about the terrain patch that lies in the cell.
Note that any cell with unknown traversability should be marked, e.g., with a negative number.

Although it is up to the user to populate the map cell values based on the associated speed distribution, it is important that the values provide interpretability and allow the user to easily adjust the riskiness of the planner. To this end, we propose to use the convex combination of the mean and $\text{CVaR}_{\alpha}(\cdot)$ at level $\alpha$ of the underlying speed distribution:
\begin{align}
m_{k,h,w} = \beta \cdot s_{k,h,w}^{\text{risk}} + (1-\beta) \cdot s_{k,h,w}^{\text{mean}}, \quad \beta\in[0,1],
\end{align}
where we know $0\leq s_{k,h,w}^{\text{risk}} \leq s_{k,h,w}^{\text{mean}}\leq \smax$. Intuitively, $m_{k,h,w}$ can be interpreted as the \textit{risk-adjusted speed} which lies between the CVaR speed estimate and the mean speed estimate due to convex combination.

For convenience, we define a look-up function $\slookup{\setM, \mathbf{p}, s}$ that returns the risk-adjusted speed estimate in the multi-layer speed map $\setM$ given a position $\mathbf{p}\in\R^2$ and speed $s\in\R$:
\begin{equation}
\slookup{\setM, \mathbf{p}, s} =\begin{cases}
0 & \text{if out-of-map or unknown},\\
m_{k,h,w} & \text{otherwise},
\end{cases}
\end{equation}
where $(k,h,w)$ are the map indices corresponding to position $\mathbf{p}$ and speed $s$ if $\mathbf{p}$ lies in the map. Note that $0$ speed is returned when the look-up is not valid, which is desirable when the unknown region can be dangerous, such as water. However, if domain knowledge suggests that the unknown is benign, a more optimistic value can be returned such as the query speed $s$.

%% file: min_time_navigation.tex
\section{Minimum Time Navigation}

\begin{figure}[t]
	\centering
	\includegraphics[width=\linewidth, trim={0cm 0cm 0cm 0cm},clip]{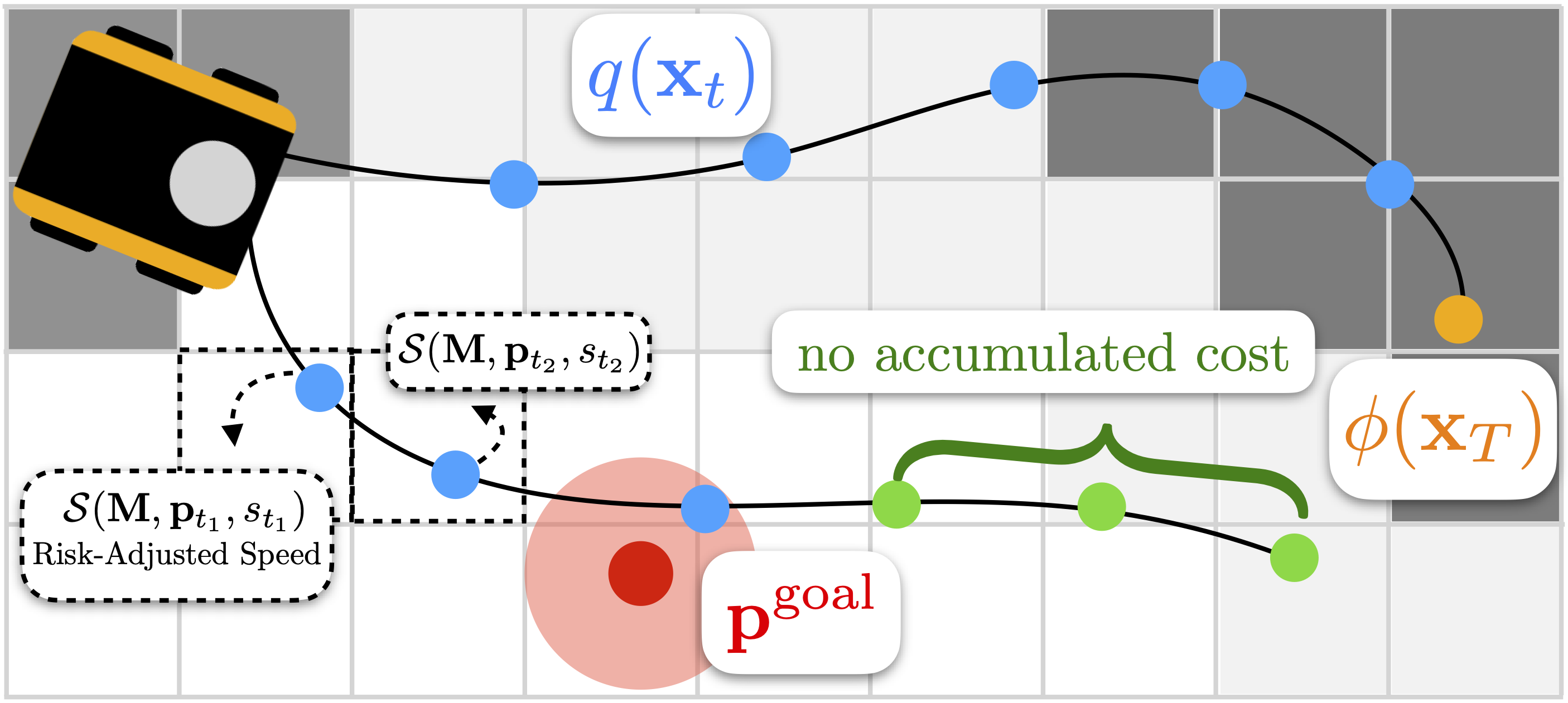}
	\caption{
		Diagram showing how each MPPI rollout is assigned cost in order to minimize time-to-goal, where the underlying colored grid is the risk-adjusted speed map. The cost~\eqref{eq:min_time} consists of the stage cost $q(\mathbf{x}_t)$ (blue) for $0\leq t\leq T-1$ and the estimated time-to-goal $\phi(\mathbf{x}_T)$ (orange). The core idea is that the nominal traversal time at each step $t$ will be adjusted based on the risk-adjusted speed estimate $\mathcal{S}(\setM, \mathbf{p}_t, s_t)$, where $s$ is speed map, $\mathbf{p}_t$ and $s_t$ are the position and speed extracted from state $\mathbf{x}_t$. 
		Note that if any state on a rollout satisfies the goal tolerance (red), the subsequent states do not accumulate stage cost or the terminal cost (shown in green).
	}
	\label{fig:mppi_and_speed_map}
\end{figure}

We adopt the MPPI controller proposed in~\cite[Algorithm~2]{williams2017information} for minimum time planning. MPPI is an information theoretic model predictive control (MPC) algorithm that tries to approximate the mean of the optimal control distribution via weighted samples of a Gaussian proposal distribution in order to minimize the KL-divergence between the two.
This approach is attractive because it is derivative-free and works with general cost functions and dynamics. Next, we follow the notation in~\cite{williams2017information} and propose a new cost function for min-time planning using the proposed risk-aware traversability map. Fig.~\ref{fig:mppi_and_speed_map} illustrates a high-level overview of our strategy and the notation used.

Consider the discrete time stochastic system:
\begin{equation}\label{eq:dynamics}
    \mathbf{x}_{t+1} = F(\mathbf{x}_t, \mathbf{v}_t),
\end{equation}
where $\mathbf{x}_t\in\R^n$ is the state vector and $\mathbf{v}_t\in\R^m\sim\mathcal{N}(\mathbf{u}_t,\mathbf{\Sigma})$ is the noisy realization of the nominal control input $\mathbf{u}_t\in\R^m$. Given the initial condition $\mathbf{x}_0$, a sequence of input $\mathbf{v}_{0:T-1}$ leads to the state trajectory $\mathbf{x}_{0:T}$ according to the dynamics~\eqref{eq:dynamics}. For the purpose of min-time planning, we assume it is possible to extract the planar position $\mathbf{p}_t\in\R^2$ and speed $s_t\in\R$ from state $\mathbf{x}_t$. Furthermore, let $\done{\mathbf{x}_{0:t}}$ be an indicator function that returns $1$ when any state $\mathbf{x}_{\tau}$ has reached the goal at position $\mathbf{p}^{\text{goal}}$ for $0\leq\tau\leq T$, and returns $0$ otherwise.

The min-time state-dependent objective is defined as
\begin{equation}\label{eq:min_time}
    C(\mathbf{x}_{0:T}) = \phi(\mathbf{x}_T)+\sum_{t=0}^{T-1} q(\mathbf{x}_t),
\end{equation}
where $\phi(\mathbf{x}_T)$ and $q(\mathbf{x}_t)$ are the time-to-go and the stage cost, respectively:
\begin{align}
\phi(\mathbf{x}_T) &= \frac{\norm{\mathbf{p}^{\text{goal}} - \mathbf{p}_T }}{s^{\text{default}}}\left( 1-\done{\mathbf{x}_{0:T}} \right) \\
q(\mathbf{x}_t) &= \frac{ s_t \cdot \Delta }{ \slookup{\setM, \mathbf{p}_t, s_t} } \left( 1-\done{\mathbf{x}_{0:t}} \right),
\end{align}
with $s^{\text{default}}$ being the default speed for estimating time-to-go at the end of the rollout, and $\Delta$ being the sampling duration. Intuitively, the lower the risk-adjusted speed $\slookup{\setM, \mathbf{p}_t, s_t}$ is, the more nominal stage cost $\Delta$ is scaled up, indicating longer time to travel. If any state $\mathbf{x}_\tau$ reaches the goal for $0\leq\tau\leq T$, all subsequent states do not incur stage cost or cost-to-go. 

At each time $t$ and given the nominal control sequence $\mathbf{u}_{t:t+T}$, MPPI estimates the mean of the optimal control distribution as the weighted sum of control rollouts that are sampled from $\mathcal{N}(\mathbf{u}_\tau,\mathbf{\Sigma})$ for all $t\leq \tau\leq t+T$. The weight of each rollout is an exponentiated cost function~\eqref{eq:min_time} evaluated along the induced state trajectory. The algorithm runs in an receding horizon fashion, where the nominal control sequence $\mathbf{u}_{t+1:t+T+1}$ in the next round is set to be newest estimate of the mean of the optimal control distribution.

%% file: results.tex
\section{Results}

In an off-road environment, where the geometric and semantic properties are hard to assess, traversability can be highly non-Gaussian. For instance, if terrain is classified as vegetation but there is no distinction between dense bushes or soft grass, the speed outcome may exhibit bi-modal distribution (i.e., being stuck or not).
With this core issue in mind, we first validate our approach of capturing speed distributions in a grid world (\cref{sec:results:gridworld}), and then integrate the proposed planner into a full autonomy stack in a high-fidelity Unity environment (\cref{sec:results:autonomy_stack}).
Given this environment and autonomy stack, \cref{sec:results:dataset} describes the process of collecting a training dataset and training a neural network to predict PMFs of the speed distribution.
Then, \cref{sec:results:phx_mppi} demonstrates the improved navigation performance as a result of the learned speed maps and risk-awareness.

\subsection{Grid World Navigation}\label{sec:results:gridworld}

\begin{figure}[t]
	\centering
	\includegraphics[width=\linewidth, trim={0cm 0cm 0cm 0cm},clip]{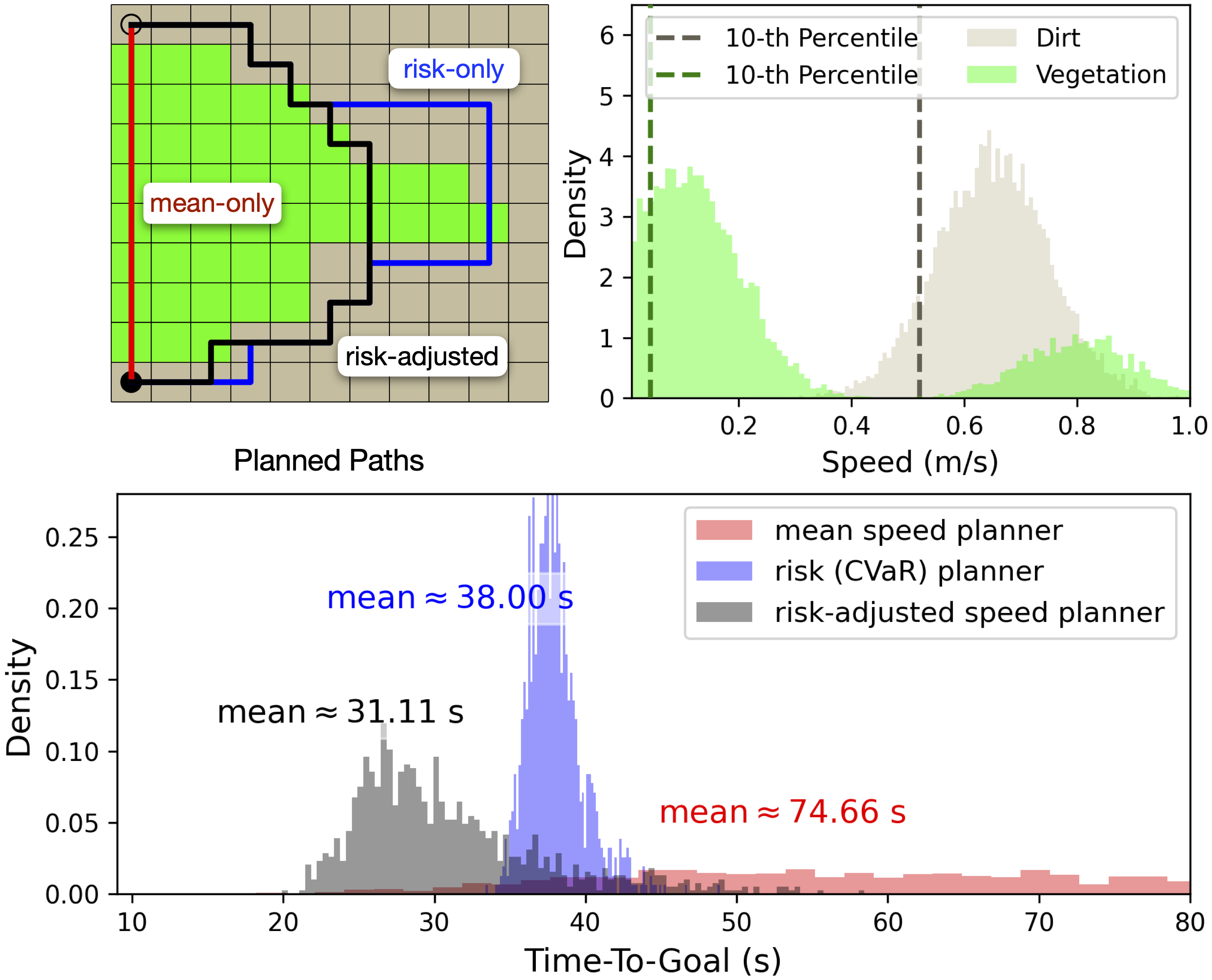}
	\caption{
		A grid world example (top left) where the green \textit{vegetation} and brown \textit{dirt} terrain types have distinct speed distributions (top right) given the fixed commanded speed. A min-time planner that ignores risk (red) has the worst time-to-goal with high variance. On the other hand, the planner that only uses CVaR speed estimates acts conservatively by avoiding most of the vegetation cells, which leads to low-variance performance but still slow in time-to-goal. By combining both the risk and mean speeds, our proposed planner (gray) gains better average performance by taking moderate risks.
	}
	\label{fig:gridworld}
\end{figure}

To validate that a speed distribution representation of traversability can be incorporated into a risk-aware planner and lead to improved performance, we first designed a grid world (see upper left of \cref{fig:gridworld}), where each $1~\text{m}\times 1~\text{m}$ cell is associated with a semantic type (either vegetation or dirt). The task is to navigate from the start position to the goal position in minimum time by planning a sequence of actions chosen from $\{ \code{UP}, \code{DOWN}, \code{LEFT}, \code{RIGHT} \}$ that move the robot to a  neighboring cell in the corresponding directions with nominal speed of $1~\text{m/s}$. Although every action is deterministic, the actual traversal time is stochastic due to the underlying speed distribution of traversed cell, as shown in \cref{fig:gridworld} (upper right). Note that we use a single-layer traversability map (i.e., $K=1$) and denote $m_{h,w}$ for each value.

To find the min-time trajectory, we use a best-first search algorithm~\cite{russell2002artificial} with a prioritized search queue where new nodes with reachable states given actions are added. The stage cost of the map cell at row $h$ and column $w$ is the estimated traversal time $1/m_{h,w}$, where CVaR is fixed at level $\alpha=0.1$. We compare the performance of the planner with $\beta\in\{\textcolor{red}{0}, \textcolor{darkgray}{0.5}, \textcolor{blue}{1}\}$, which correspond to using mean speed, risk-adjusted speed, and the (pure) risk speed. The optimal trajectories and their average time-to-goal values over $1000$ trials are visualized in \cref{fig:gridworld} (upper left and bottom). By only capturing the mean speeds (\textcolor{red}{$\beta=0$}), the planner's performance has high variance because it does not consider the worst-case. By only accounting for risk (\textcolor{blue}{$\beta=1$}), the planner's performance has very low variance, but it is also highly conservative. By utilizing both risk and mean of the speed distribution (\textcolor{darkgray}{$\beta\in(0,1)$}), the planner's riskiness can be chosen to achieve shorter average time-to-goal with slight increase in variance. 
This experiment demonstrates that representing traversability as a speed distribution can be incorporated into a risk-aware planner and lead to improved performance.

\subsection{Integration with Autonomy Stack in High-Fidelity Environment}\label{sec:results:autonomy_stack}

Next, we integrate the proposed methods into a full autonomy stack~\cite{arlautonomystack} in a high-fidelity Unity simulation environment with a Clearpath Warthog platform~\cite{clearpathwarthog}, as shown in~\cref{fig:unity_and_test_goals}.
To focus on the challenge of coarse semantic labeling and its impact on traversability analysis, the Unity environment contains \textit{dirt} and \textit{vegetation} terrain types, where roughly $\frac{1}{4}$ of the bushes (classified as \textit{vegetation}) on grass are non-traversable (i.e., the robot cannot drive through them), which slows down navigation and the resulting wheel slip has a side-effect of causing substantial drift in the vehicle's pose estimate.

\subsection{Data Collection and Network Training}\label{sec:results:dataset}
\begin{figure}[t]
	\centering
	\includegraphics[width=\linewidth, trim={0cm 0cm 0cm 0cm},clip]{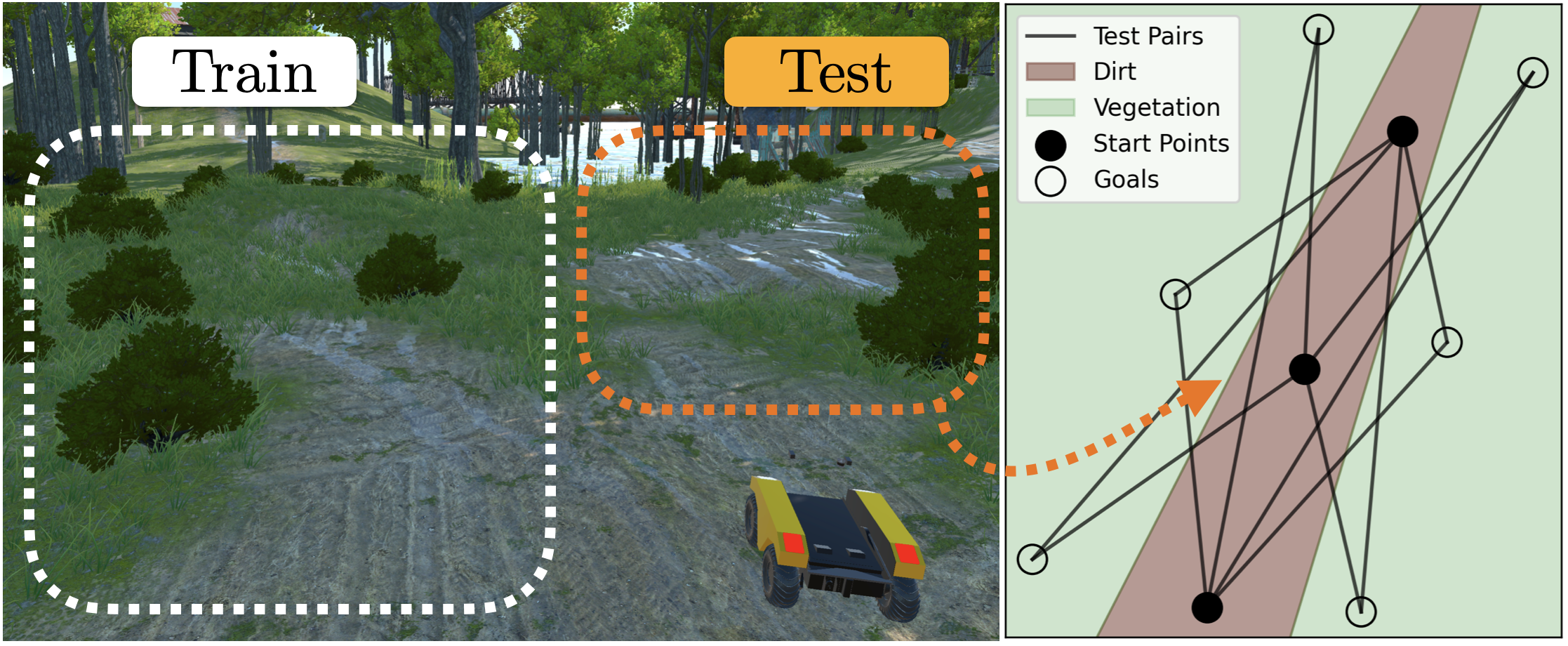}
	\caption{
	    Unity environment (left) where the robot learns and uses speed distribution maps for min-time planning. The key challenge is that bushes ($1/4$ of which lead to collision) are not distinguished semantically from the collision-free grass. After learning the distributions in the training region, the robot is tested to reach goals specified in regions with vegetation (right). 
	}
	\label{fig:unity_and_test_goals}
\end{figure}

\begin{figure}[t]
	\centering
	\includegraphics[width=\linewidth, trim={0cm 0cm 0cm 0cm},clip]{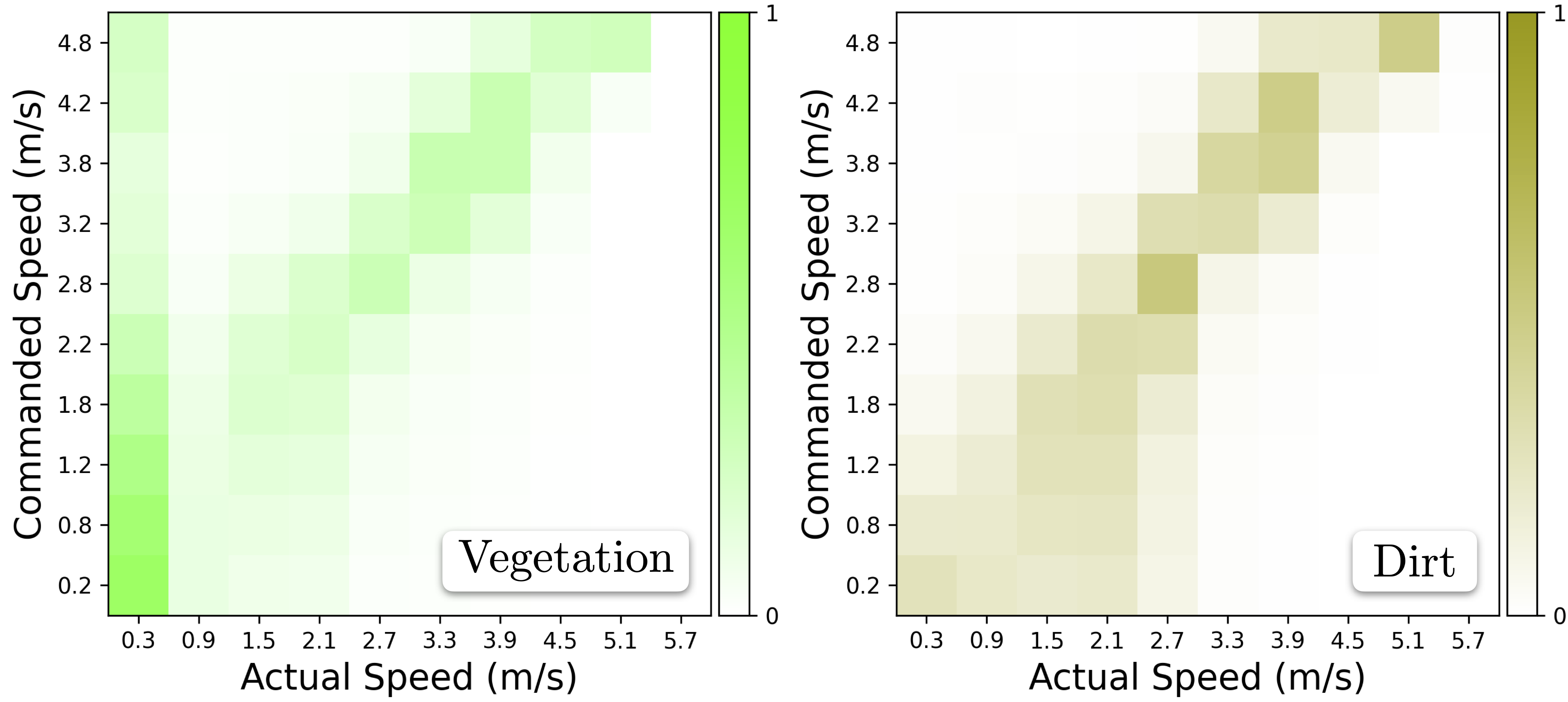}
	\caption{
		Learned speed distributions for vegetation and dirt terrains, where the color gradient indicates magnitude of the probability mass. Note the bi-modal distribution associated with vegetation, whose outcome speeds are concentrated around $0$~m/s (due to collisions with bushes) and around the commanded speeds. On the other hand, driving on dirt has a uni-modal, roughly linear relationship between commanded speed and realized speed.
	}
	\label{fig:learned_sdm}
\end{figure}

In order to learn the traversability model as defined in~\eqref{eq:cond_distribution}, samples consisting of tuples of (\textit{commanded speed}, \textit{terrain type}, \textit{realized speed}) were gathered with a joystick-controlled robot, where the joystick provided the commanded speeds, and the traversed terrain type and true speed were taken directly from the Unity simulation engine. The robot drove for $3$ minutes in the training area (Fig.~\ref{fig:unity_and_test_goals} left), resulting in about $7000$ and $2000$ samples associated with the vegetation and dirt terrain, respectively. The dataset was used to train a multi-layer feedforward NN with $2$ hidden layers of $64$ nodes and ReLU activation functions. The network input contains the one-hot encoding of the terrain type and the commanded speed, and the output consists of the probability mass function (PMF) for $10$ output speed bins between $0$~m/s and maximum speed $5$~m/s via a softmax layer.
The network was trained for $10$ epochs with the Adam optimizer with the learning rate of $0.005$, resulting in the speed distribution maps visualized in \cref{fig:learned_sdm}, where the rows correspond to $10$ binned commanded speeds and their resultant PMFs of speed outcomes.

During deployment, top-down $100\times 100$ semantic images of the environment with $0.4$~m cell resolution was processed by the network, as illustrated in \cref{fig:architecture}. The training and testing areas contain mostly dirt and vegetation terrains types, and unknown semantic types were assumed to induce $0$~m/s. Note that the semantic map and every commanded speed are paired and reshaped into a large batch input to the network. A $10$ layer $100\times 100$ speed distribution map with $10$ output speed bins can be evaluated under $2$~ms using the CPU (all runtimes reported on a desktop computer with an Intel i7-7700K CPU and 32GB RAM).
The risk-aware speed maps can be extracted from the speed distribution maps as the convex combination of PMF mean and CVaR. When the entire autonomy stack was running and competing for CPU, mean and CVaR took $15$~ms and $150$~ms to compute, respectively. The computation for CVaR was done via rectangular approximation of density within each output speed bin (software optimization and GPU parallelization could likely reduce the CVaR calculation times substantially). Due to computational constraints, the traversability map was published at $2$~Hz to the MPPI local planner.

\subsection{Min-Time Navigation Benchmark}\label{sec:results:phx_mppi}
In order to benchmark the effectiveness of learned risk-aware speed maps, the planner is tasked to navigate the robot from a set of pre-specified starting positions to goal positions in an unseen test environment, as illustrated in~\cref{fig:unity_and_test_goals}. Each test pair of starting point and goal were repeated for 3 trials over a range of risk weights $\beta\in[0.0, 0.15, 0.3, 0.45, 0.6]$. The goal tolerance was set to be a $3$~m circle and the longest distance the robot had to travel was about $35$~m.  A timeout period of $40$~s was imposed to terminate the trials where the robot got stuck or disoriented due to collisions with bushes. 
The planner models the Clearpath Warthog robot as a differential drive robot whose control input consists of two wheel speeds. During each MPPI optimization round, $500$ control rollouts were sampled over $5$~s horizon at $20$~Hz according to noise standard deviation of $5$~rad/s for each wheel.
The estimated optimal control distribution was iteratively refined as MPPI ran in a receding horizon fashion. A small default speed of $0.5$~m/s was used for estimating time-to-go to encourage the robot to approach the goal.

\begin{figure}[t]
	\centering
	\includegraphics[width=0.9\linewidth, trim={0cm 0.3cm 0cm 0cm},clip]{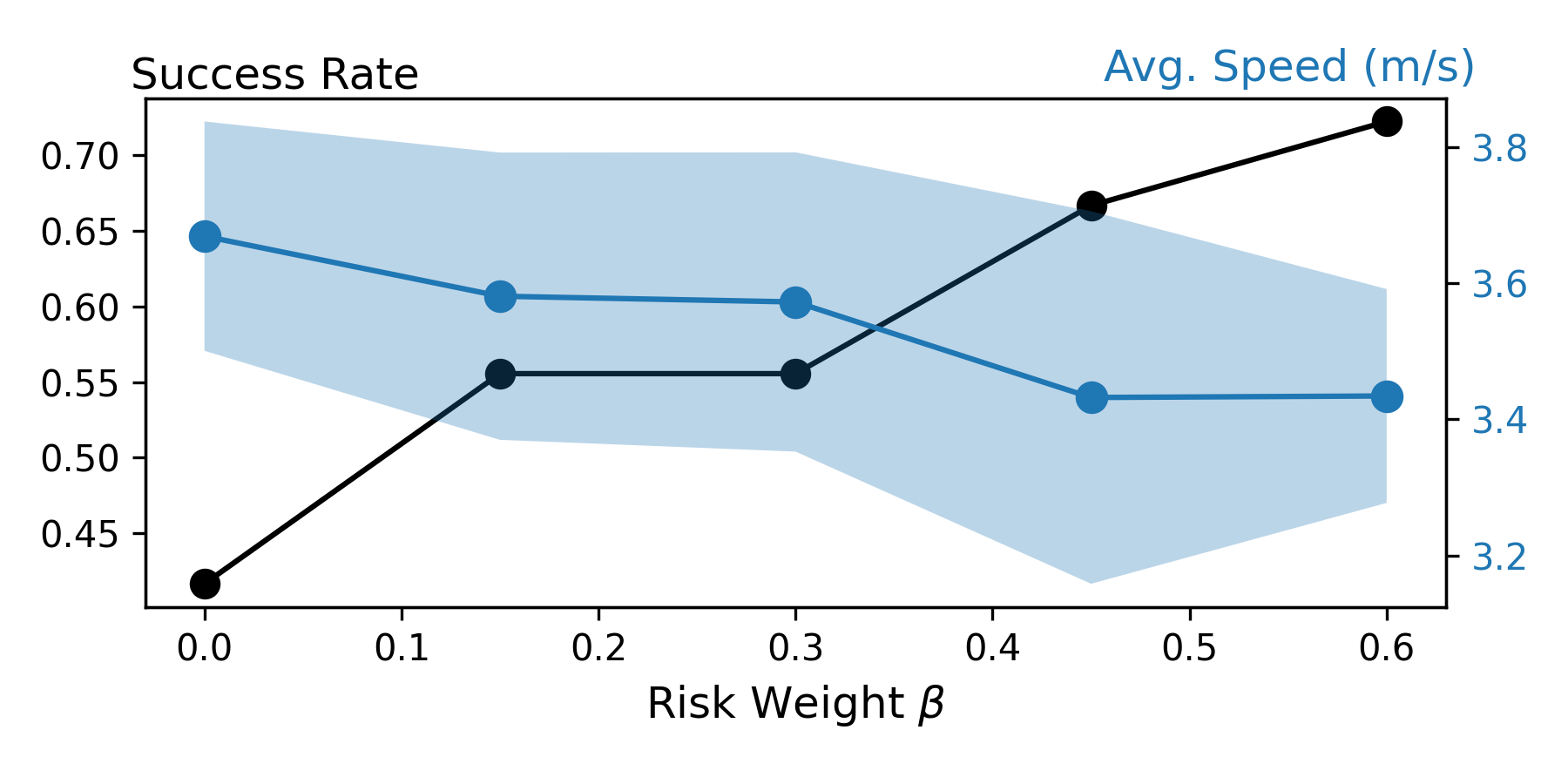}
	\caption{
		Effect of risk weight $\beta\in[0,1]$ on average speed of successful trials and the overall success rate, where CVaR is fixed at level $\alpha=0.1$. By increasing $\beta$ from $0$ to $0.6$, the robot achieves a much higher success rate ($42\% \to 72\%$), while the average speed reduces moderately as a performance trade-off. However, very high risk weight (e.g., $\beta > 0.8$) can cause excessive conservatism.
	}
	\label{fig:benchmark}
\end{figure}

\begin{figure}[t]
	\centering
	\includegraphics[width=0.65\columnwidth, trim={0cm 0cm 0cm 0cm},clip]{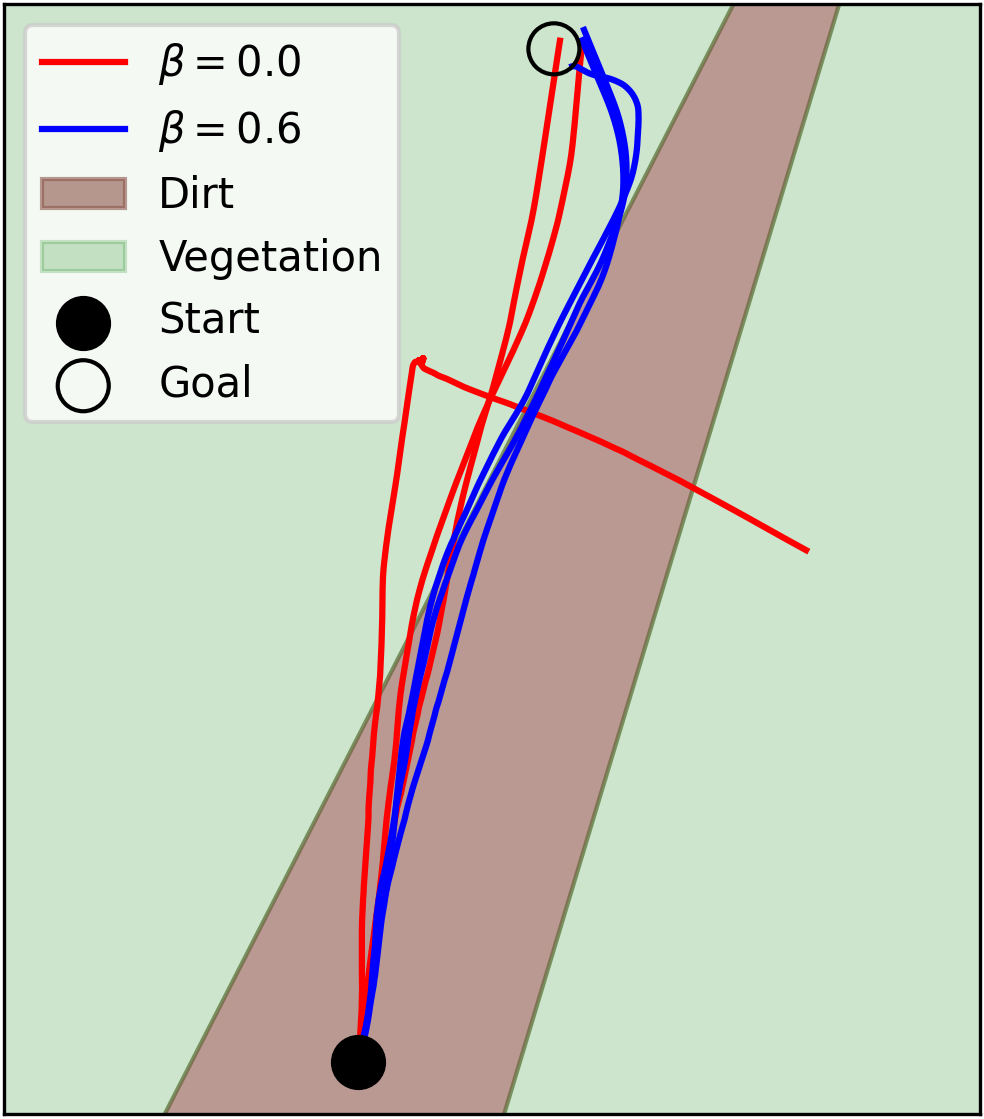}
	\caption{
		Example MPPI min-time trajectories with risky ($\textcolor{red}{\beta=0}$) and conservative ($\textcolor{blue}{\beta=0.6}$) behaviors. The risky planner's trajectories (red) follow a relatively straight path to the goal by traversing longer distance on vegetation, but there's a higher risk of collisions with bushes. Collision did happen to one of the trials and led to odometry failure due to wheel slips, which made the robot move in the wrong direction. On the other hand, the risk-aware planner's trajectories spend longer time on dirt, thus are more likely to reach the goal safely.
	}
	\label{fig:mppi_example}
\end{figure}

The benchmark results are shown in~\cref{fig:benchmark} which contain the success rate and the average speeds over the successful trials for a range of risk weights. As the risk weight increases, the robot has lower average speed but higher success rate ($42\%\rightarrow 72\%$) as beta increases from $0$ to $0.6$. Example rollouts produced by $\textcolor{red}{\beta=0}$ and $\textcolor{blue}{\beta=0.6}$ are shown in~\cref{fig:mppi_example}, where the risk-aware trajectories ($\textcolor{blue}{\beta=0.6}$) overlap more with the dirt terrain (lower risk), whereas the planner that only accounts for the expectation ($\textcolor{red}{\beta=0}$) prefers shorter paths that overlap more with vegetation (higher risk). As a result, one of the trajectories led to a collision with bushes, which caused localization errors and failure to reach goal.

%% file: conclusion.tex
\section{Conclusion \& Future Work}
This work proposed a new notion of traversability as the conditional speed distribution achievable by a robot, conditioned on the environment semantics and commanded speed. This representation can be learned directly from experienced trajectories and can be incorporated into various planning paradigms as a speed map. The proposed planning strategy was shown to lead to faster average time-to-goals compared to other methods that did not consider the worst-case. Lastly, the proposed risk-aware strategy led to higher success rate in minimum-time navigation task in a high-fidelity simulator. 

One area of future work is in automatically tuning the risk parameter $\beta$ online, based on differences between realized speed and commanded speed. 
Additionally, the work could be extended to capture other forms of uncertainty, such as from out-of-distribution inputs or from probabilistic outputs from the semantic segmentation module.
Finally, a learned cost-to-go estimator could be used to improve the cost assigned to the end of each MPPI rollout for improved performance.